\documentclass[letterpaper, 10 pt, conference]{ieeeconf}  
\usepackage{amsmath,amsfonts}
\usepackage{color}
\usepackage{array}
\usepackage[caption=false,font=normalsize,labelfont=sf,labelformat=simple,textfont=sf]{subfig}
\usepackage{textcomp}
\usepackage{stfloats}
\usepackage{url}
\usepackage{verbatim}
\usepackage{graphicx}
\usepackage{cite}
\usepackage{amsmath,amssymb}
\usepackage{makecell}
\usepackage{bbding}
\usepackage{multirow} 
\usepackage{multicol} 
\usepackage{arydshln}
\usepackage{booktabs}

\usepackage{graphicx}
\usepackage{float}
\usepackage{svg}  
\usepackage{balance}

\UseRawInputEncoding
\IEEEoverridecommandlockouts                              

\pdfoutput=1
\title{\LARGE \bf
S\&Reg: End-to-End Learning-Based Model for Multi-Goal Path Planning Problem
}

\author{Yuan Huang$^{1}$, Kairui Gu$^{1}$, and Hee-hyol Lee$^{1}$
\thanks{$^{1}$Yuan Huang, Kairui Gu, and Hee-hyol Lee are with the Graduate School of Information, Production and Systems, Waseda University, Kitakyushu, Japan
        {\tt\small gakki@toki.waseda.jp, gkrcarry@ruri.waseda.jp, hlee@waseda.jp}}%
}

\begin{document}

\maketitle
\thispagestyle{empty}
\pagestyle{empty}

\begin{abstract}

In this paper, we propose a novel end-to-end approach for solving the multi-goal path planning problem in obstacle environments. Our proposed model, called S\&Reg, integrates multi-task learning networks with a TSP solver and a path planner to quickly compute a closed and feasible path visiting all goals. Specifically, the model first predicts promising regions that potentially contain the optimal paths connecting two goals as a segmentation task. Simultaneously, estimations for pairwise distances between goals are conducted as a regression task by the neural networks, while the results construct a symmetric weight matrix for the TSP solver. Leveraging the TSP result, the path planner efficiently explores feasible paths guided by promising regions. We extensively evaluate the S\&Reg model through simulations and compare it with the other sampling-based algorithms. The results demonstrate that our proposed model achieves superior performance in respect of computation time and solution cost, making it an effective solution for multi-goal path planning in obstacle environments. The proposed approach has the potential to be extended to other sampling-based algorithms for multi-goal path planning.

\end{abstract}

\section{Introduction}

Multi-goal path planning problem is a fundamental task for robots that aims to find a closed and feasible path to visit a sequence of goals, which is required for various applications, including data collection\cite{ref1,ref2,ref3}, pickup-and-delivery \cite{ref4,ref5}, and manufacturing \cite{ref6}. In addition to feasibility, the length of the path is a crucial criterion for evaluating the cost of the solution, particularly for industrial robots \cite{ref7} and fly vehicles \cite{ref3,ref8}. Generally, the multi-goal path planning can be decomposed into two subproblems: Travelling Salesman Problem (TSP) and a path-finding problem. The TSP determines the visiting order for goals, while a path planner connects the goals referring to the order. 

\begin{figure}[t]
	\centering
	\includegraphics[width=3.5 in]{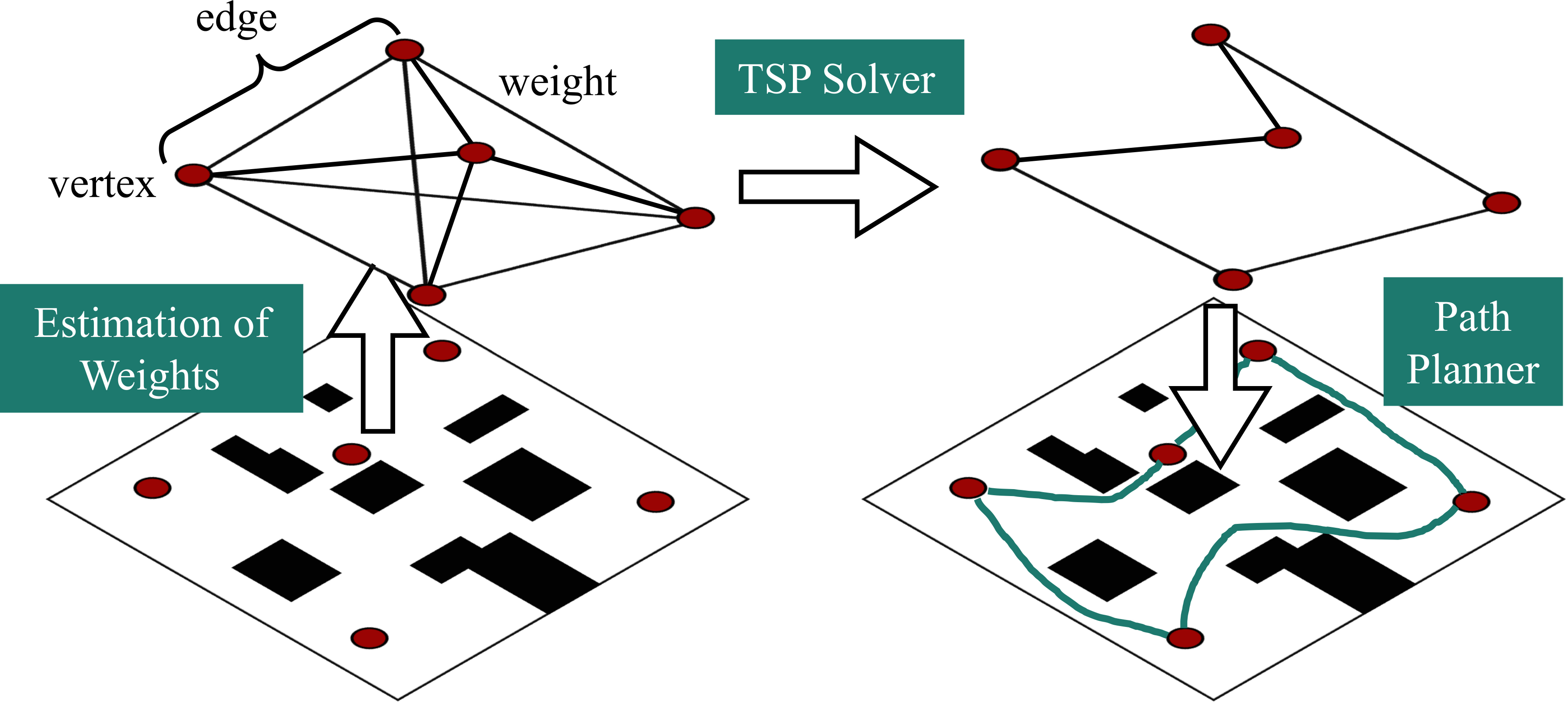}

	\caption{Illustration of the multi-goal path planning problem. Black and white areas represent obstacles and feasible configuration space. As a physical TSP, goals as well as vertices are dented by red nodes. An edge represents a connection path between pairwise goals. Weights are estimated as lengths of the connected paths before a TSP solver and a path planner.}
	\label{fig1}
\end{figure}

However, traditional TSP solvers \cite{ref9,ref10,ref11} have limited effectiveness in solving the ordering problem in obstacle environments, as they are typically designed for unconstrained settings. In particular, the physical TSP \cite{ref12}, which accounts for the TSP in obstacle environments, returns an order utilizing an Euclidean distance to compute a weight between two vertices as shown in Fig. \ref{fig1}. Nevertheless, the Euclidean distance cannot precisely represent the true distance between vertices in obstacle environments, which results in improper orders with increased costs. 

Since the traditional methods cannot meet the efficiency requirement, sampling-based path planning algorithms are utilized to estimate the distance between vertices. Rapidly-exploring Random Tree Star (RRT*) \cite{ref13} is a representative sampling-based algorithm, which constructs an exploring tree in the configuration space to search for an optimal path incrementally, ensuring the accuracy of the path length as the weight for TSP. However, RRT* grows by the guidance of random samples, which theoretically results in massive computation, especially in complex environments with narrow passages. Additionally, it is computationally expensive to build a complete weight graph with $O(M^2)$ complexity for $M$ goals, which is necessary to ensure the optimality of the visiting order. Alternative approaches have been introduced, such as the multi-tree extension strategy discussed in \cite{ref8} that refrains from using a symbolic task planner, and the space-filling forest \cite{ref14,ref15} consisting of multiple trees until the forest fills the full configuration space. In \cite{ref16}, Chandak develops and extends the informed sampling strategy by sampling in selected regions and de-emphasizing regions for multiple goals. Nonetheless, they suffer from the shortcomings of the uniform sampling strategy, resulting in redundant samples and exploration. Moreover, the growth of the multiple trees or forest is terminated even though some edges (paths) between vertices (goals) have not been established. Consequently, the exploration returns an asymmetric weight matrix for the TSP solver, which probably leads to a non-optimal order. Therefore, there is a need for new approaches that can more effectively address the multi-goal path planning problem in obstacle environments and break these restrictions.

In this paper, we focus on enhancing the performance in respect of calculation time and solution cost of the sampling-based algorithms to address the multi-goal path planning problem via a learning-based model 
(including a \textbf{Seg}mentation task and a \textbf{Reg}ression task), called S\&Reg. We design multi-task learning networks to estimate the pairwise distance between goals for further computing the least-cost order via a TSP solver. Simultaneously, a region that surely contains the optimal path connecting pairwise goals is predicted by networks, which performs as a heuristic to conduct the subsequent exploration of a sampling-based path planner. The S\&Reg is experimentally compared with the other sampling-based algorithms in respect of calculation time and solution cost in different scenarios. 

Our main contributions are three-fold:
\begin{enumerate}
	\item{Designing an end-to-end learning-based model, S\&Reg, for the multi-goal path planning problem, which is available to integrate with other sampling-based algorithms.}
	\item{Devising multi-task learning networks, including a segmentation task for predicting promising regions and a regression task for estimating the distance between pairwise goals.}
	\item{Extensive simulations verifying the remarkable performance of the S\&Reg in both simple and complex environments with different numbers of goals.}\\
\end{enumerate}

The rest of the paper is organized as follows. In Section II, we formulate the multi-goal path planning problem and briefly introduce the sampling-based algorithms. The proposed S\&Reg model with the multi-task learning networks is presented in Section III. In Section IV, simulations are conducted with other sampling-based algorithms. Finally, Section V presents our conclusion, highlighting the key contributions and potential directions for future research.
\section{Preliminaries}

\subsection{Problem Definition}

Multi-goal path planning problem can be defined as a task to find a closed and feasible path visiting a sequence of goals with minimum cost. Let $\mathcal{C}$ be the configuration space, and $\mathcal{C}_f$ denotes the feasible space for a robot. $\mathcal{C}_o$ is a set of obstacles, while $\mathcal{C}_o=\mathcal{C}-\mathcal{C}_f$. Generalized as a physical Travelling Salesman Problem (TSP), an undirected complete graph $\mathcal{G}(\mathcal{V},\mathcal{E})$ with accurate weights $\mathcal{W}$ is significant to compute the goal sequence for visiting in obstacle environments. Here, $\mathcal{V}$ is a set including $M$ goals, $\mathcal{V}=(v_1,\cdots,v_M)$, $v_i \in \mathcal{C}_f$. Edge $\mathcal{E}=(e_{i,j}): i,j\in[1,\cdots,M]$ is a set that includes all possible connections between any two vertices. The weight $w_{i,j}$  on an edge represents the cost of the path between two vertices, $v_i$ and $v_j$. We use $\Sigma$ as an optimal visiting order. Thus, the multi-goal path planning problem can be defined as:

$$
 \Sigma=arg  \min  (\sum_{i=1}^{M}\sum_{j=1}^{M}x_{i,j}*w_{i,j}) \eqno{(1)} 
$$
$$
s.t.\ \forall i \in [1,M],\sum_{j=1}^{M}x_{i,j}=1 \eqno{(2)}
$$
$$
\forall j \in [1,M],\sum_{i=1}^{M}x_{i,j}=1 \eqno{(3)}
$$
$$
\forall i,j \in [1,M],x_{i,j}=
\begin{cases}
	1 & \text{if a path from}\  v_i\, \text{to}\, v_j \,\text{is visited}\\
	0 & \text{otherwise}
\end{cases}  \eqno{(4)}
$$
$$
w_{i,j}	\approx c(\sigma_{i,j}) \eqno{(5)}
$$(2) and (3) guarantee that all vertices are visited. Especially, for the multi-goal path planning problem, let $\sigma_{i,j}(t)$ be a feasible path connecting vertices $v_i$ and $v_j$. In (5), the weight on the edge is denoted by a local cost $c(\sigma_{i,j})=\int_{0}^{1}\|\sigma_{i,j}(t)\|dt$ , where $\sigma_{i,j} (0)=v_i$, and $\sigma_{i,j} (1)=v_j$. It should be noted that it is difficult to estimate the loca cost without any information of the local path. Thereby, some local path planers are employed to explore the local paths for estimating the local cost.
\subsection{Sampling Strategy}
 \begin{figure}[tb]
	\centering
	\includegraphics[width=3.3 in]{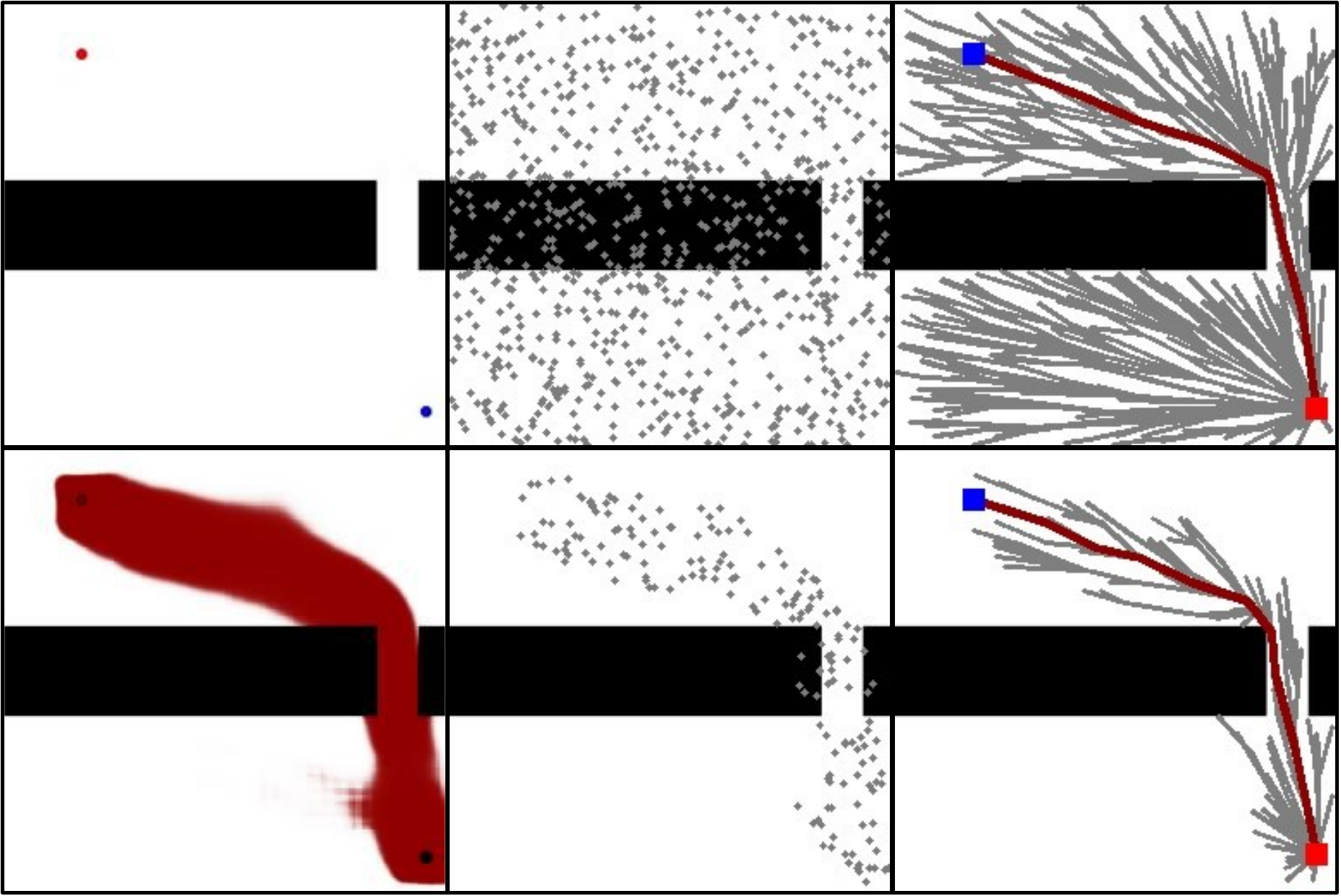}
	
	\caption{Comparison between uniform and heuristic sampling strategies. Exploration trees are growing by gray samples to connect the goals. A predicted region in dark red guides the search with promising samples.}
	\label{fig2}
\end{figure}
Sampling-based algorithms such as RRT \cite{ref17}, PRM \cite{ref18}, and their variants, RRT*, and PRM*\cite{ref13} are widely used as path planners for the multi-goal path planning. The efficiency of these algorithms heavily relies on the quality of samples. Generally, a uniform sampling strategy is employed to explore the environment, allowing the search to theoretically reach any position. However, redundant samples may be generated during the asymptotically optimal process of solutions. The major direction of improvements is to reduce the sampling domain for non-uniform sampling, as introduced in \cite{ref19,ref20,ref21}. Recently, a new direction for enhancing the sampling strategy via learning-based models is introduced in \cite{ref22}. As illustrated in the second row of Fig. \ref{fig2}, a promising region is generated by learning-based networks, where the optimal path is probably located. The promising region is deployed as heuristics to realize the non-uniform sampling and reduce the exploration time. It is evident that the heuristic sampling strategy requires far fewer samples to find the optimal path compared with the uniform sampling strategy.

\section{S\&Reg Model}

\begin{figure*}[t!]
	\centering
	\includegraphics[width=7 in]{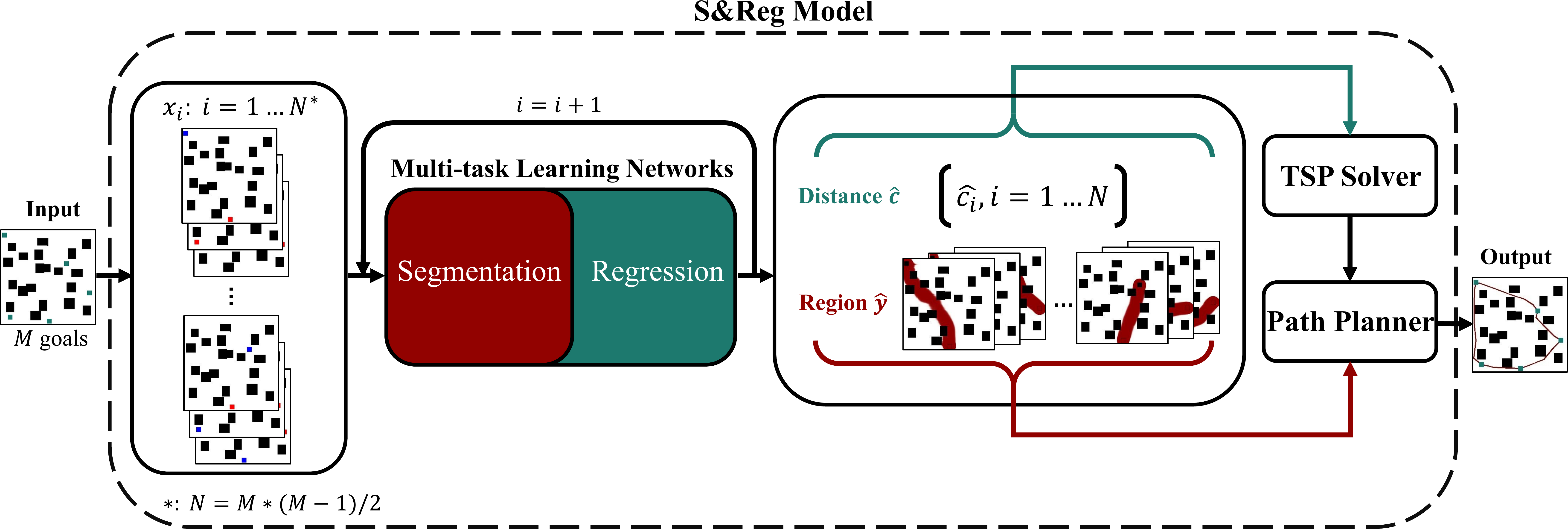}
	
	\caption{Illustration of S\&Reg Model (Segmentation and Regression). }
	\label{fig3}
\end{figure*}

In this section, we outline the S\&Reg model, which contains multi-task learning networks, a TSP solver, and a path planner. As illustrated in Fig. \ref{fig3}, an input map with $M$ goals is divided into $N=M*(M-1)/2$ sub-maps (denoted as $x_i:i \in [1\cdots N]$) with pairs of goals. Then the sub-maps are fed into the multi-task learning networks alternately. The outputs of the networks are subsequently transferred to the TSP solver and path planner to compute the final solution.
 \begin{figure}[t!]
	\centering
	\includegraphics[width=3.4 in]{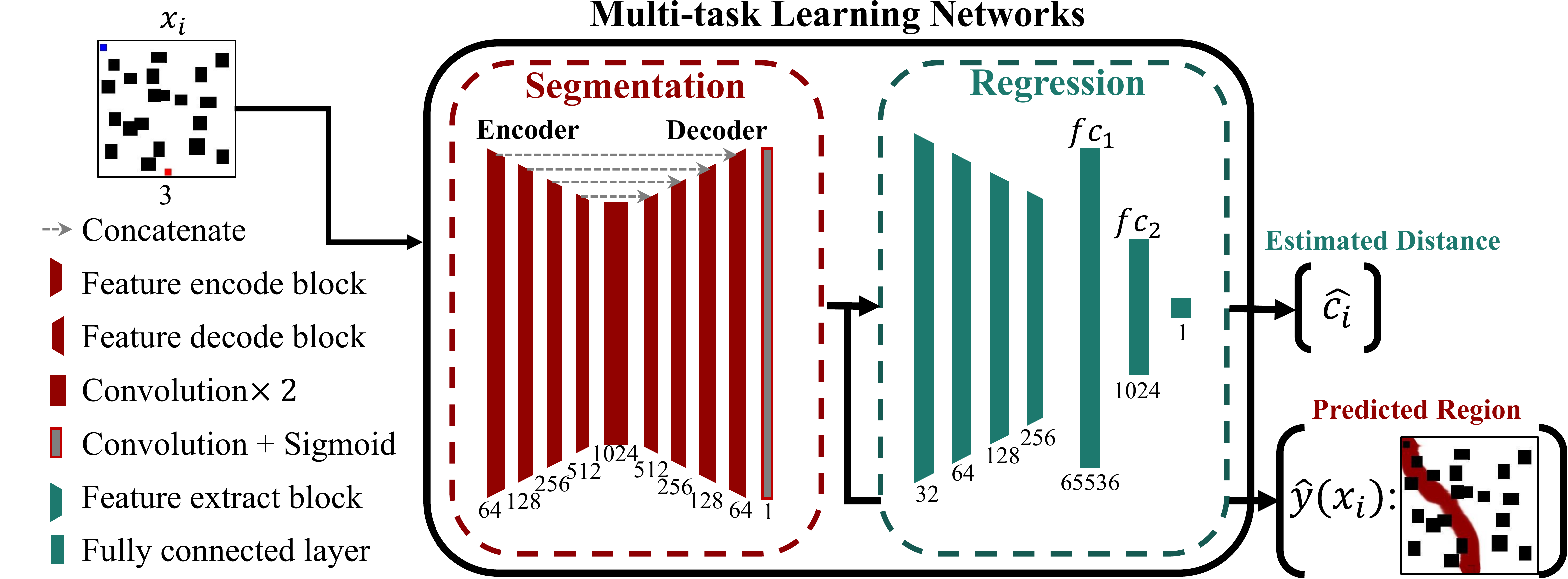}
	
	\caption{Structure of the multi-task learning networks. Numerals represent the dimensions of feature maps.}
	\label{fig4}
\end{figure}
\subsection{Multi-Task Learning Networks}

\begin{figure}[t!]
	\centering
	\includegraphics[width=3.4 in]{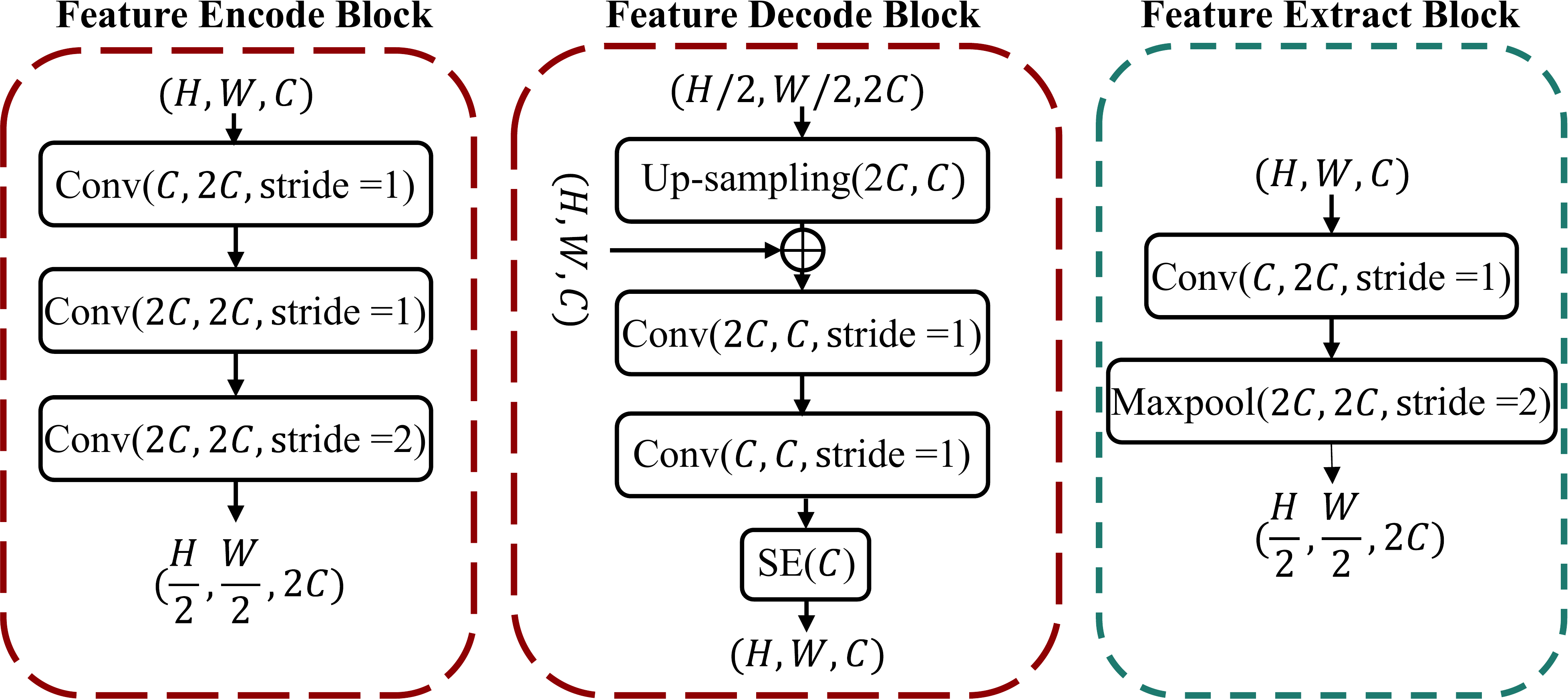}
	
	\caption{Basic blocks for the segmentation task and the regression task in the multi-task learning networks. The shape and channel number of the feature map are denoted by $H$, $W$, and $C$, respectively.}
	\label{fig5}
\end{figure}

We design multi-task learning networks to predict a promising region and estimate the distance between pairwise goals, which is fed up with RGB maps $x_i$ split from the original map. The structure of the multi-task learning networks is depicted in Fig. \ref{fig4}.\\
\textbf{Segmentation Task:} We employ an encoder-decoder architecture to extract deep-level and shallow-level features simultaneously. The encoder comprises feature encoder blocks (as shown in Fig. \ref{fig4}) to shrink down the input shape and capture contextual information. To alleviate the information loss during down-sampling, a 2-stride convolutional layer is equipped. Symmetrically, the decoder utilizes feature decode blocks as depicted in Fig. \ref{fig5}. The concatenate operation is adopted from the UNet \cite{ref23}, while inputs of the feature decode block are aggregated with the corresponding layers in encode. To enhance the fusion efficiency between the contextual feature maps, a Squeeze-and-Excitation \cite{ref24} (SE) layer is embedded before an up-sampling operation. As the shape of the feature map matches the shape of the input, a plain convolutional layer followed by a sigmoid activation function performs to produce the predicted region (denoted as $\hat y$). The predicted region, as the ground truth is shown in white in Fig. \ref{fig6}(b), denotes areas with a high likelihood of providing promising samples, while black areas indicate the opposite.\\
\textbf{Regression Task:} We transform the estimation of the distance between goals into a one-dimensional output regression task. Based on the predicted region, a basic convolutional operation with 1 stride performs to extract features, while a $MaxPool$ operation pools the convolution results by a factor of 2 in both dimensions. As shown in Fig. \ref{fig4}, this feature extract block reduces the size of the feature map from (256,256,1) to (16,16,256), which is then flattened into a single vector with 65536 neurons, denoted as $fc_1$. The extracted features are combined using another fully connected layer (denoted as $fc_2$) with 1024 neurons, and a single output neuron is used to output the estimated distance $\hat c_i$ for the regression task.\\
\textbf{Weighted Geometric Loss:} To ensure that both tasks are learned efficiently, we utilize a geometric loss in our multi-task learning networks. This approach aims to balance the individual losses by reducing the differences in gradient and scale. Additionally, we incorporate a weighting strategy inspired by the Focused Loss Strategy (FLS) \cite{ref25} to prioritize more important tasks and losses. For the segmentation task, we combine a Binary Cross Entropy (BCE) loss with a Dice loss to evaluate the results at pixel-level and map-wise. The formulations are as follows:
$$ 
\mathcal{L}_{1}(y,\hat y)=-\sum_{i=1}^{H}\sum_{j=1}^{W}y_{i,j}*\log(\hat{y}_{i,j}) \quad\quad\quad\quad\quad
$$
$$ 
\quad\quad\quad\quad\quad\quad\quad\quad\quad\quad\quad\quad+(1-y_{i,j})*\log(1-\hat{y}_{i,j}),
\eqno{(6)}
$$
$$\mathcal{L}_{2}=1-\frac{2|y\cap\hat{y}|}{|y|+|\hat{y}|},\eqno{(7)}
$$
$$|y\cap\hat{y}|=\sum_{i=1}^{H}\sum_{j=1}^{W}y_{i,j}*\hat{y}_{i,j},\eqno{(8)}
$$
$$|y|+|\hat{y}|=\sum_{i=1}^{H}\sum_{j=1}^{W}y_{i,j}*y_{i,j}+\sum_{i=1}^{H}\sum_{j=1}^{W}\hat{y}_{i,j}*\hat{y}_{i,j}.\eqno{(9)}
$$$H$ and $W$ are the size of the input map, while $y$ and $\hat{y}$ represent the labeled region and predicted region for input $x$, respectively. For the regression task, a Mean Square Error (MSE) loss is utilized to evaluate the difference between the estimated distance $c_i$ and the true distance $\hat{c}_i$ by 
$$ \mathcal{L}_{3}=\frac{1}{N}\sum_{i=1}^{N}(c_i-\hat{c}_i)^{2}.
\eqno{(10)}
$$Finally, the total loss is defined as:
$$ \mathcal{L}_{Total}=\sum_{i=1}^{T}\alpha_i*log(\mathcal{L}_i),
\eqno{(11)}
$$where $T$ represents the number of loss functions, and $\alpha_i$ is the importance for each loss.
\subsection{TSP Solver}
As depicted in Fig. \ref{fig2}, the output of the regression task can be interpreted as an estimated distance between each pair of goals. After the $M$ times prediction, a complete graph is implemented for TSP. We adopt Elkai \cite{ref10} solver based on  Lin-Kernighan Heuristics (LKH)\cite{ref26} method to address TSP, which is known to yield the optimal solution based on the weight matrix and determine the visiting order for the multi-goal path planning problem.

\subsection{Path Planner}

To enhance the efficiency of the pathfinding process, we adopt the predicted region as a heuristic to generate promising samples in the region. As a traditional sampling-based algorithm, RRT explores the configuration space by expanding the branches of a tree. Since the optimal solution is surely lying in the region, the samples in the region are promising to improve the current solution than randomly generated samples. A hybrid sampling based on a heuristic sampler and a goal-biased sampler is designed as follows:
$$
NewSample = \begin{cases}
	H_{sampler} & \text{if}\ rand() > k\\
	GB_{sampler} & \text{otherwise}
\end{cases}, \eqno{(12)}
$$where a heuristic-biased coefficient $k$ is preset. The heuristic sampler $H_{sampler}$ generates samples in the promising region randomly. Goal-biased sampler $GB_{sampler}$ allows the exploration to be biased towards the goal for guiding the search. The exploration terminates when a feasible path is found to connect two goals.

	\begin{figure} [t]
	\footnotesize	
	\centering
	\setcounter {subfigure} {0} (a){
		\includegraphics[width=3.2    in]{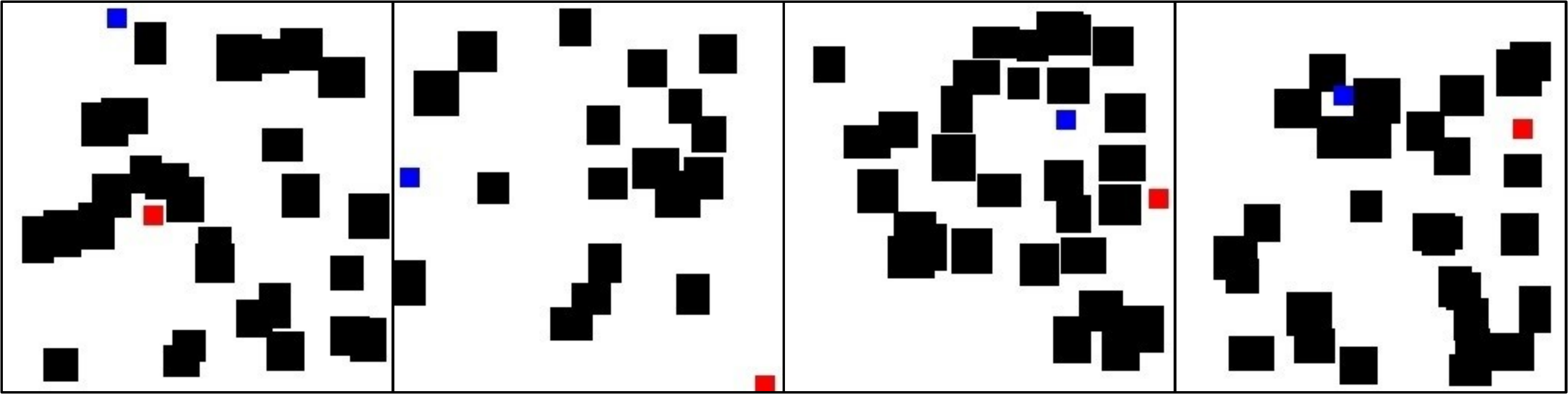}
		\label{fig6:a}}\\
	\setcounter {subfigure} {1} (b){
		\includegraphics[width=3.2    in]{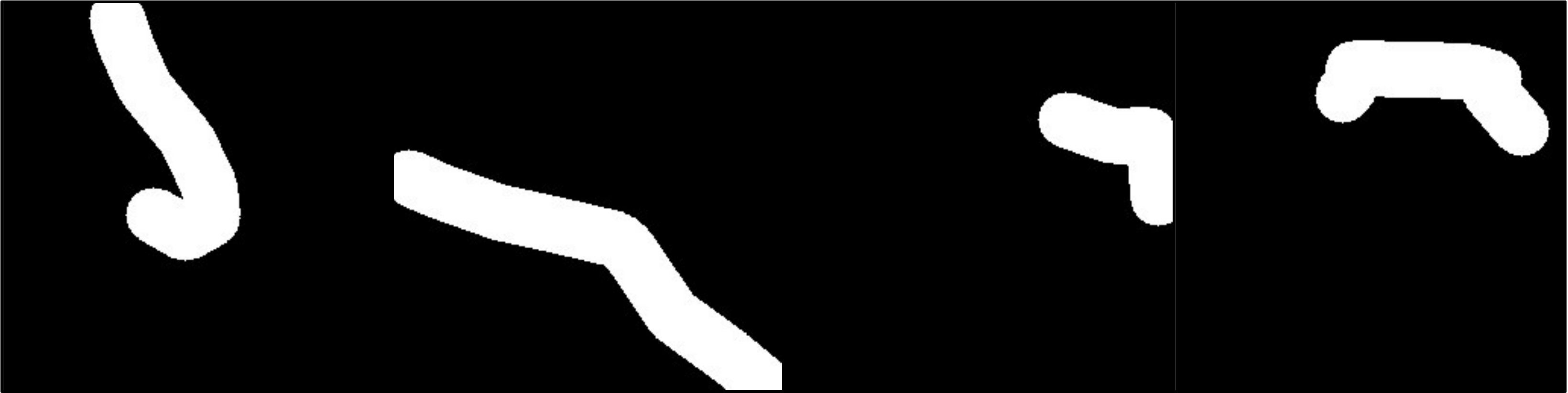} \label{fig6:b}}\\
	\setcounter {subfigure} {3} (c){
		\includegraphics[width=3.2    in]{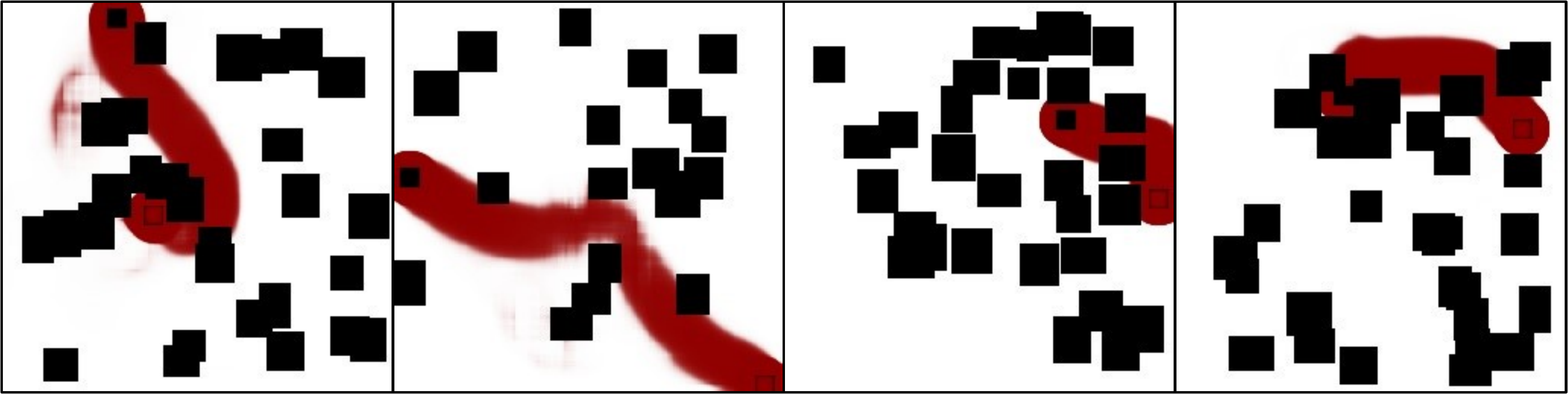} \label{fig6:d}}\\

		\caption{Illustration of training examples in the dataset MGPFD and predicted regions by the S\&Reg. (a) Example maps in the training dataset. (b) Labeled Regions. (c) Predicted regions of the test dataset.}
	\label{fig6} 
\end{figure}
\section{Simulation \& Results}

In this work, we establish a dataset for multi-goal path planning. Also, training details are introduced in this section. Besides, simulations are carried out to verify the efficiency of the S\&Reg model both in calculation time and solution cost in different scenarios. 

\subsection{Training Details and Results of Multi-Task Learning Networks}

The proposed Multi-Goal Path Finding Dataset (MGPFD) contains 16000 RGB maps with two goals and obstacles. It is available at https://github.com/RTPWDSDM/MGPFD. As shown in Fig. \ref{fig6}(a), the example map size is 256$\times$256, while two goals are denoted in blue and red, respectively. Randomly generated black obstacles are placed throughout the map. Each ground truth in the dataset includes a labeled region that is dilated from the optimal path, and a corresponding length, which represents the true distance between the goals.
 \begin{figure}[t!]
	\centering
	\includegraphics[width=3.4 in]{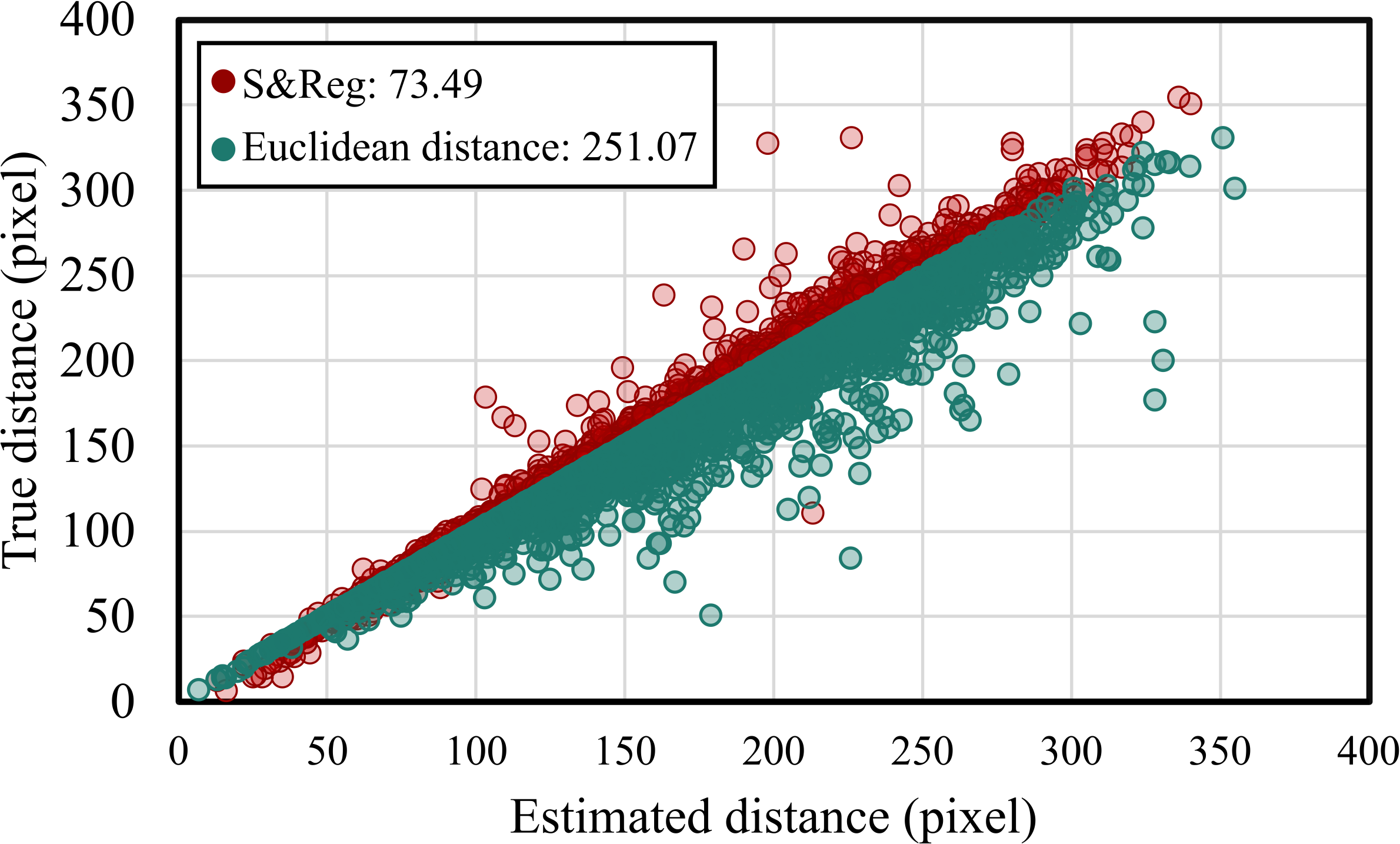}
	
	\caption{Distribution of the estimated and true distance with MSE results using the test dataset.}
	\label{fig7}
\end{figure}
\begin{table}[t]
	\caption{Ordering Results of the Sampling-Based Algorithms for 20 Times Running}
	\label{table_example}
	\begin{center}
		
		\begin{tabular}{ccc}
			
			\hline
			& \makecell[c]{Simple scenario\\ (5 goals)} & \makecell[c]{Complex scenario \\(12 goals)}\\
			\hline
			RRT* & \makecell[c]{\textbf{0,1,2,3,4,0}\\ 0,3,2,1,4,0\\
				0,2,1,3,4,0} &\makecell[c]{0,1,4,2,8,9,3,7,10,6,11,5,0\\
				0,1,5,11,6,10,7,3,9,8,2,4,0\\
				0,1,8,9,3,2,4,11,7,10,6,5,0\\
				0,1,8,9,3,7,10,6,5,11,4,2,0\\
				\textbf{0,1,8,9,3,7,10,6,11,5,4,2,0}\\
				0,1,8,9,3,7,10,6,11,5,2,4,0\\
				0,1,8,9,10,7,3,2,4,11,6,5,0}\\
			\hline
			Euclidean RRT* & \textbf{0,1,2,3,4,0}&
			\textbf{0,1,8,9,3,2,4,7,10,6,11,5,0}\\
			\hline
			SFF & \textbf{0,1,2,3,4,0} &\makecell[c]{0,1,3,9,8,2,4,11,7,10,6,5,0\\
				0,1,4,2,8,9,3,7,10,6,11,5,0\\
				0,1,6,10,7,3,9,8,2,4,11,5,0\\
				0,1,8,2,4,11,6,10,7,3,9,5,0\\
				\textbf{0,1,8,9,3,2,4,11,7,10,6,5,0}\\
				0,1,8,9,3,7,10,6,5,11,4,2,0\\
				0,1,10,7,3,9,8,2,4,11,6,5,0\\
				0,5,6,10,7,3,9,8,1,2,4,11,0\\}\\
			\hline
			S\&Reg	& 	\textbf{0,1,2,3,4,0} &	\textbf{0,1,8,9,3,7,10,6,5,11,4,2,0}\\
			\hline
			
		\end{tabular}
	\end{center}
\end{table}

\begin{figure}[t]
	\centering
	\includegraphics[width=3.4 in]{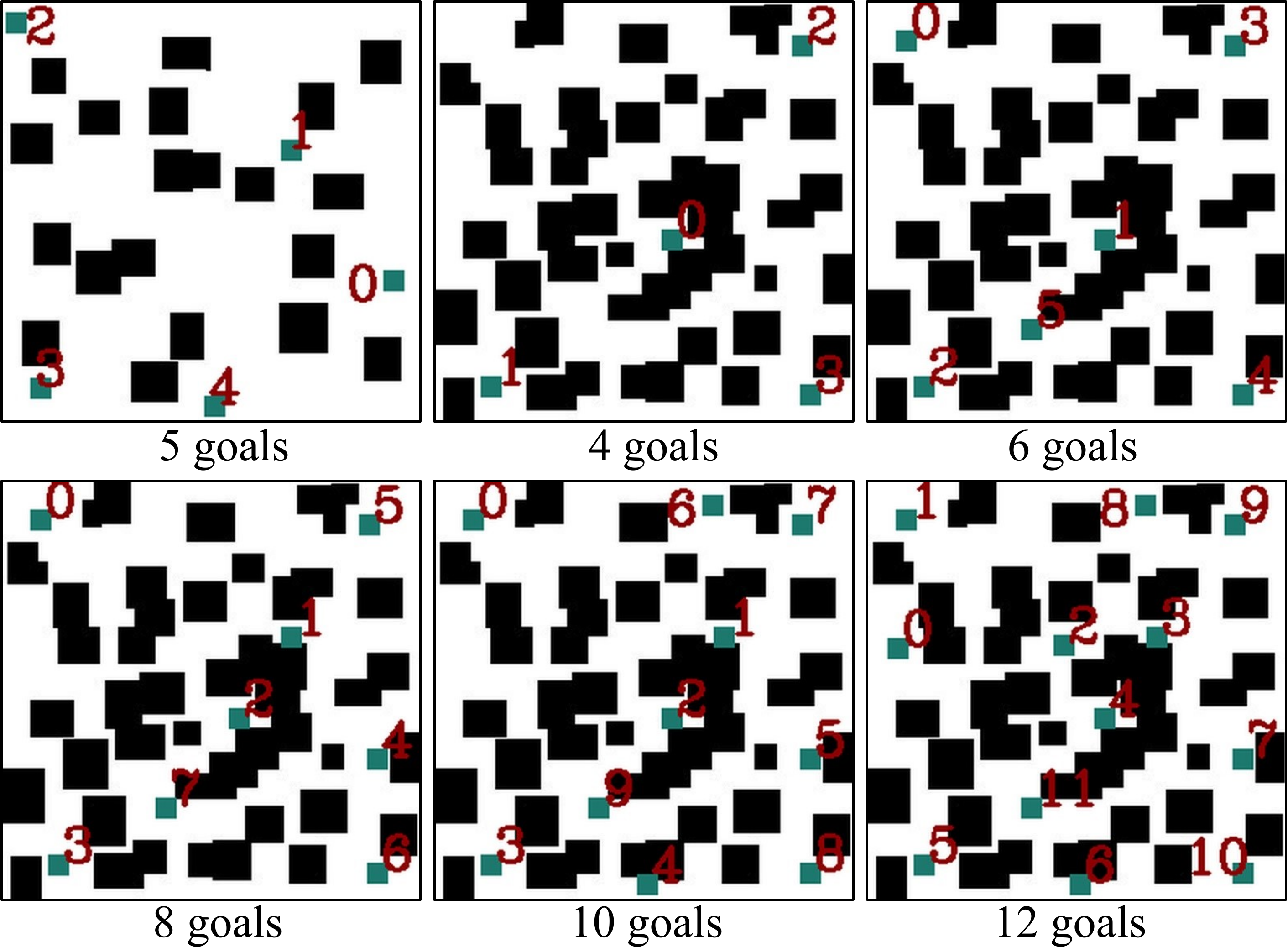}
	
	\caption{Illustration of simple and complex scenarios with different number targets.}
	\label{fig8}
\end{figure}

Our model is implemented in PyTorch using a 16GB NVIDIA GeForce 3080 Ti GPU. We train the multi-task learning networks with the Adam optimizer and set the learning rate using the Cosine Annealing Warm Restarts method \cite{ref27} with an initial learning rate of 0.0003, $T_0=1$, and $T_{mult}=2$. The dataset is split into a training dataset, a validated dataset, and a test dataset with a ratio of 6:2:2. The segmentation results are displayed in Fig. \ref{fig6}(c), with darker red indicating a higher likelihood of producing promising samples. Fig. \ref{fig7} shows the distribution of the estimated and true distances using the test dataset. The estimated distance should be linearly positively correlated with the true distance. The S\&Reg model accurately estimates the length of the optimal path between pairwise goals, as shown by the lower MSE value compared with the Euclidean distance. 

\subsection{Multi-Goal Path Planning in Different Scenarios}
To assess the performance of the S$\&$Reg, we conducted a comparative analysis of RRT*, RRT* with Euclidean distance (denoted as Euclidean RRT*), Space-filling forest (SFF), and S\&Reg in various scenarios. We ran each algorithm independently 20 times and the maximum sampling number is set to 2000. For the S\&Reg, the heuristic biased coefficient $k$ is set to 0.1. In the simple scenario, we set the goal number to five, while in the complex scenario with narrow passages, we set it to 12. Fig. \ref{fig8} shows the visual representation of the scenarios, and for each algorithm, we used the Elkai solver to determine the visiting order.

 \begin{figure}[t]
	\centering
	\includegraphics[width=3.4 in]{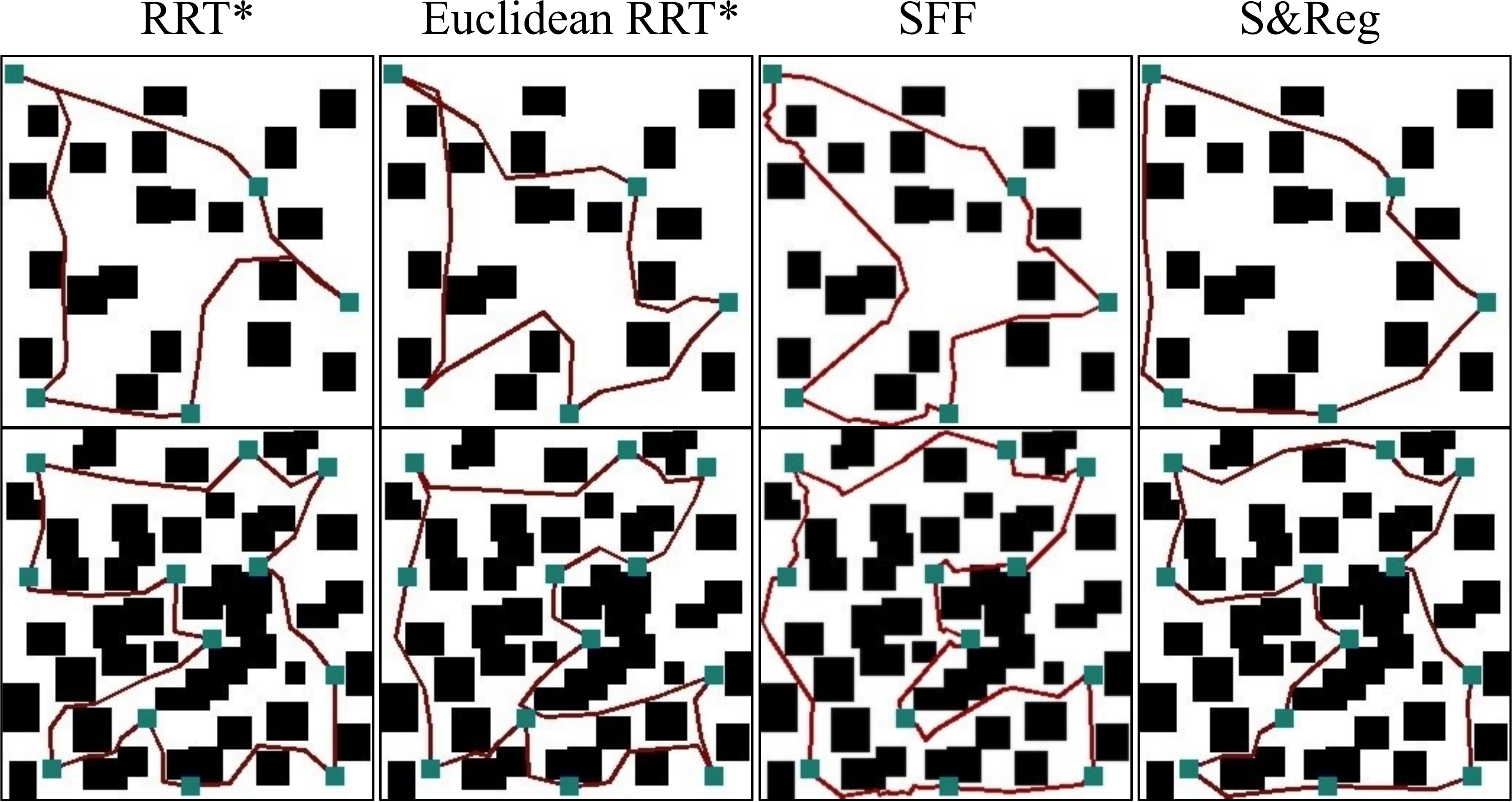}
	
	\caption{Illustration of results by RRT*, Euclidean RRT*, SFF, and S\&Reg.}
	\label{fig9}
\end{figure}
\begin{figure}[t]
	\centering
	\includegraphics[width=3.4 in]{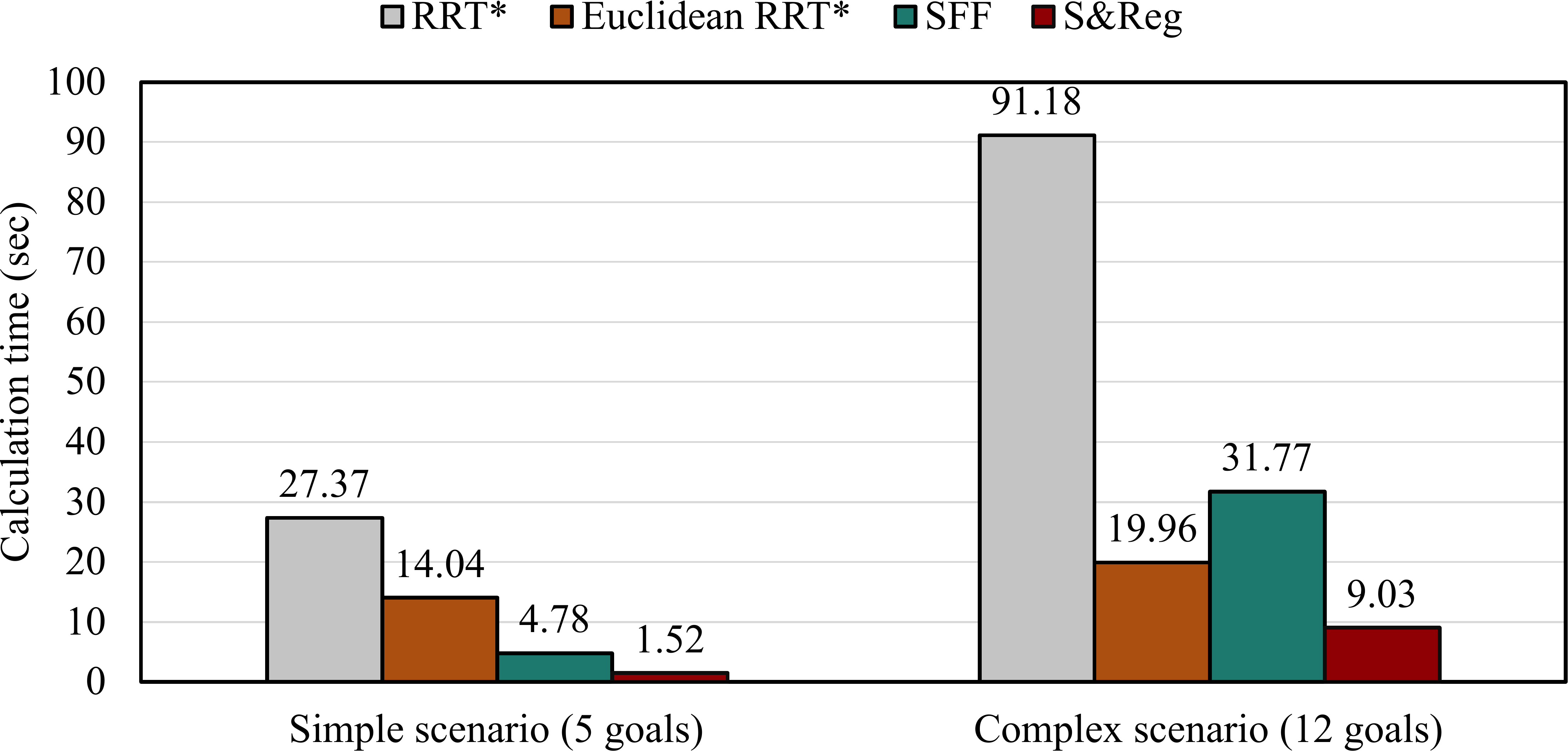}
	
	\caption{Comparison results of the solution cost in simple and complex scenarios.}
	\label{fig10}
\end{figure}

\begin{figure}[t!]
	\centering
	\includegraphics[width=3.4 in]{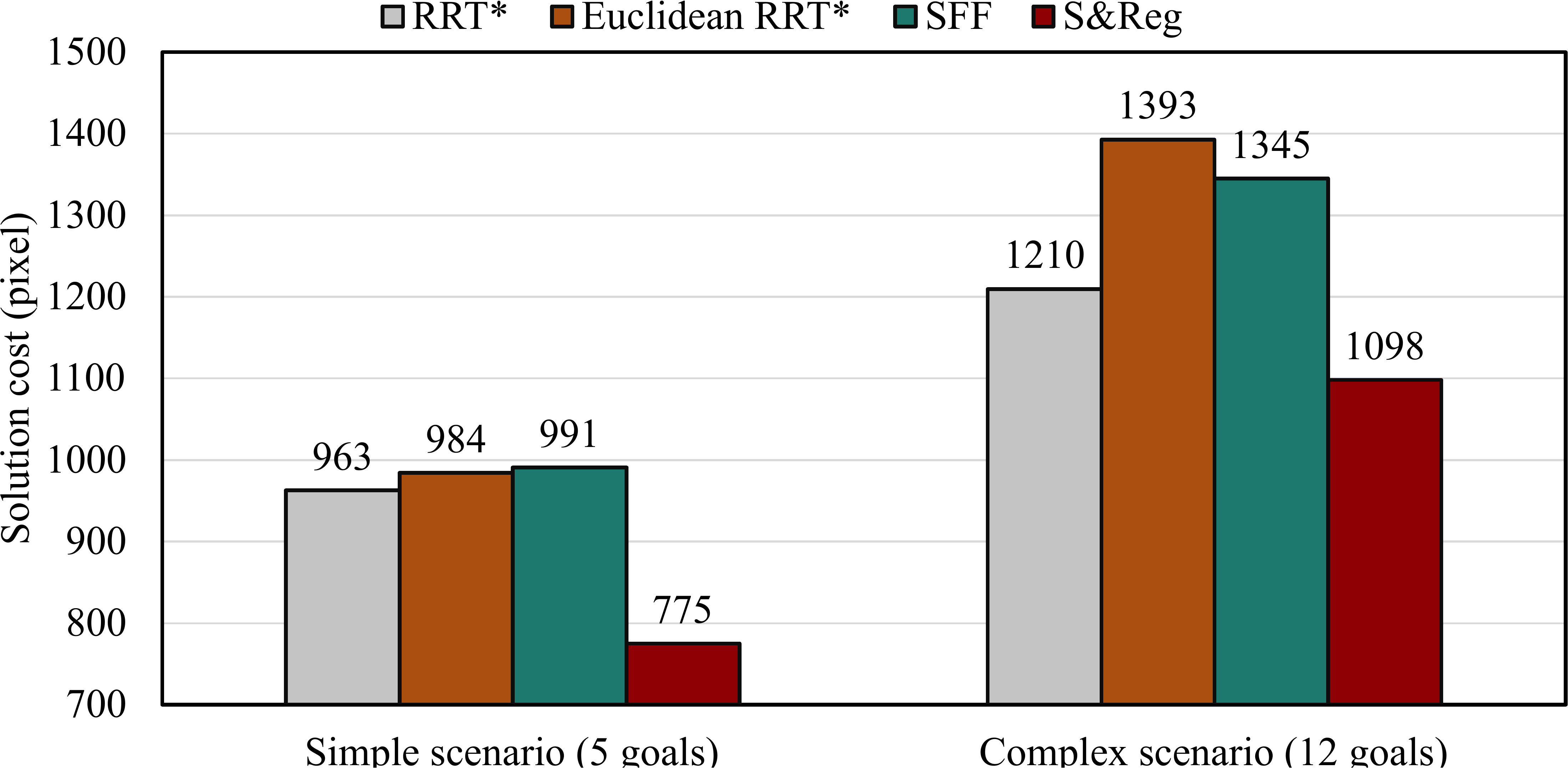}
	
	\caption{Comparison results of the calculation time in simple and complex environments.}
	\label{fig11}
\end{figure}

The TSP results of 20 times running are listed in TABLE 1, while the best results of different algorithms are highlighted in bold. In the simple scenario, all algorithms provide the same visiting order as the optimal solution but differ in the local paths between goals. Interestingly, RRT* returns alternate and non-optimal orders, resulting from inefficient exploration. In the complex scenario, which includes narrow passages and 12 goals, the TSP results of each algorithm vary considerably. RRT* and SFF have difficulty guaranteeing the identical weight matrix and ordering after each run. Additionally, The SFF suffers from the incomplete weighted graph due to its space-filling exploration forest. Euclidean RRT* and S\&Reg stably obtain the same orders, but the Euclidean RRT* fails to compute a proper weight matrix in dense obstacle environments. The final closed paths are illustrated in Fig. \ref{fig9}. The comparison results of the solution cost and calculation time are further detailed in Fig. \ref{fig10} and Fig. \ref{fig11}, respectively. Notably, the S\&Reg model reliably outperforms other multi-goal path planning algorithms in both simple and complex scenarios in terms of the final solution cost. The neural network-based distance estimation in TSP and the predicted region for the regression task accelerate the exploration. Moreover, the S\&Reg model takes much less time to tackle the multi-goal path planning problem compared with the other algorithms, as shown in Fig. \ref{fig11}.

\begin{figure}[t!]
	\centering
	\includegraphics[width=3.4 in]{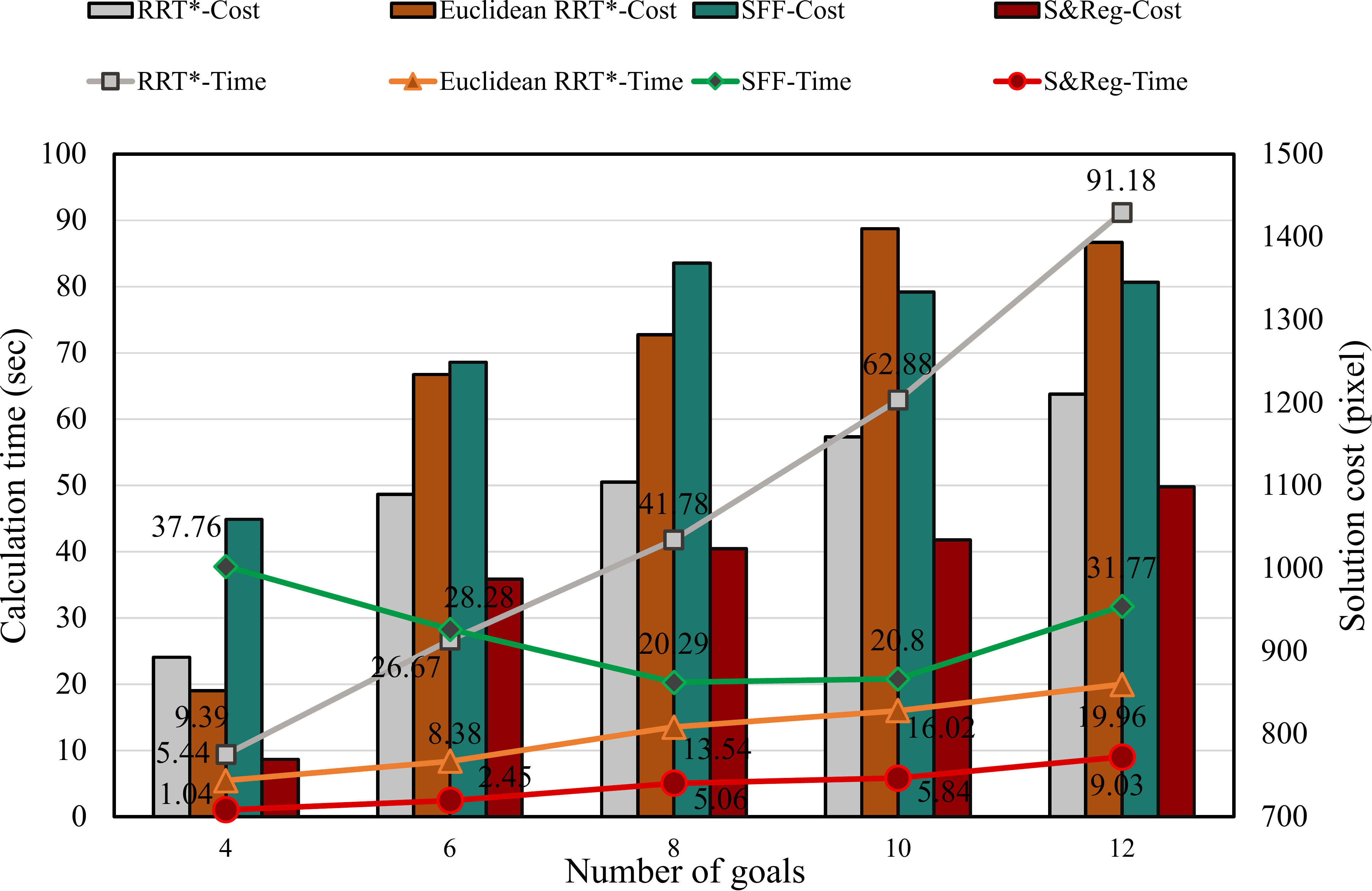}
	
	\caption{Comparison results of solution costs and calculation time with different number goal in the complex environment.}
	\label{fig12}
\end{figure}
 
\subsection{Multi-Goal Path Planning with Different Goals}
To further demonstrate the feasibility and efficiency of the S\&Reg model in complex environments, we conducted a comparison of multi-goal path planning algorithms with different numbers of goals in the complex scenario, as shown in Fig. \ref{fig8}. The results of the calculation time and solution cost are illustrated in Fig. \ref{fig12}. The S\&Reg model requires approximately 0.01 seconds for one prediction, and the time spent by the local path planner corresponds to the number of goals. It is evident that the S\&Reg model's calculation time is significantly reduced while the solution cost is remarkably improved in all situations. RRT* achieves good performance in the solution cost but sacrifices more calculation time. Furthermore, RRT* requires the number $N=M*(M-1)/2$ runs to obtain the distance matrix. Although Euclidean RRT* only requires $M$ runs of the path planner based on the Euclidean distance, its solution cost performance is almost the worst. For SFF, a forest grows to fill the entire configuration space, and the calculation time decreases if the goals can easily connect with each other during the initial growth. Besides, we attempt 24 goals to test, while the S$\&$Reg still outperforms the SFF in both calculation time (around 35 seconds) and path length. Unfortunately, the SFF takes more than 200 seconds to obtain a feasible solution. In conclusion, the S\&Reg model's results demonstrate its efficiency in both calculation time and solution cost for the multi-goal path planning problem compared with the other sampling-based algorithms.

\section{Conclusions}

In this paper, we presented an end-to-end model, named S\&Reg, to enhance the efficiency of sampling-based algorithms for the multi-goal path planning problem. To solve the physical TSP, we divided the original map into different associations of two vertices. We developed multi-task learning networks that precisely estimated the weights between two vertices and predicted a promising region for the next path planner. Specifically, we treated the weight estimation as a regression task, while the predicted region segmentation was a segmentation task aimed at identifying the optimal path's promising region. To train the model, we utilized a weighted geometric loss, enabling learning for different tasks. After prediction, we established a complete weighted graph to determine the visiting order, followed by the RRT to connect the goals efficiently, leading by the promising regions. Simulations in both simple and complex environments demonstrated the S\&Reg model's efficiency in terms of calculation time and solution cost compared with the other sampling-based algorithms. This model can be extended easily to other sampling-based algorithms for multi-goal path planning. 

\addtolength{\textheight}{0cm}   




\section*{ACKNOWLEDGMENT}

This work was supported by JST SPRING, Grant Number JPMJSP2128.


\vfill
\end{document}